\documentclass[conference]{IEEEtran}
\usepackage{times}

\usepackage[numbers]{natbib}
\usepackage{multicol}
\usepackage[bookmarks=true]{hyperref}
\usepackage{tabularx}
\usepackage{xcolor, colortbl}
\usepackage{booktabs}
\usepackage{multirow}
\usepackage{amsmath}
\usepackage{amsfonts}
\usepackage{algorithm}
\usepackage{svg}
\usepackage{algpseudocode}
\usepackage{mathtools}
\usepackage{makecell}
\usepackage{comment}
\usepackage{tikz}
\usepackage{paracol}
\usepackage{graphics}

\usepackage[utf8]{inputenc}
\usepackage{pgfplots}
\DeclareUnicodeCharacter{2212}{−}
\usepgfplotslibrary{groupplots,dateplot}
\usetikzlibrary{patterns,shapes.arrows}
\pgfplotsset{compat=newest}
\usetikzlibrary{arrows}
\usepackage{xcolor}

\usepackage{xparse}
\usepackage[normalem]{ulem}

\usepackage{amssymb}

\NewDocumentCommand\NewIndexedVar{mmm}{
\expandafter\NewDocumentCommand\csname#1\endcsname{se{^_}}{#2\IfNoValueTF{##2}{}{^{##2}}\IfNoValueTF{##3}{\IfBooleanTF{##1}{}{_#3}}{_{##3}}}
}

\NewDocumentCommand\idiff{}{t}
\NewDocumentCommand\Ndiff{}{T}
\NewDocumentCommand\itime{}{n}
\NewDocumentCommand\Ntime{}{N}
\NewDocumentCommand\itraj{}{k}
\NewDocumentCommand\Ntraj{}{K}

\NewDocumentCommand\Dset{}{\mathcal{D}}
\NewDocumentCommand\Tset{}{\mathcal{T}}
\NewDocumentCommand\traj{}{\boldsymbol{\tau}}
\NewDocumentCommand\seq{}{\boldsymbol{o}}
\NewDocumentCommand\goal{}{\boldsymbol{g}}
\NewDocumentCommand\p{mm}{#1\left(#2\right)}

\NewDocumentCommand\gpol{}{\p{\pi}{\act\middle|\state,\goal}}
\NewDocumentCommand\pol{}{\p{\pi}{\act\middle|\state}}

\NewDocumentCommand\act{}{\boldsymbol{a}}
\NewDocumentCommand\state{}{\boldsymbol{s}}

\begin{document}

\title{Goal-Conditioned Imitation Learning using Score-based Diffusion Policies}


\author{\authorblockN{Moritz Reuss, 
Maximilian Li, 
Xiaogang Jia and 
Rudolf Lioutikov} 
\authorblockA{Intuitive Robots Lab, 
Karlsruhe Institute of Technology, Germany}}


%

\maketitle

\begin{abstract}
We propose a new policy representation based on score-based diffusion models (SDMs).
We apply our new policy representation in the domain of Goal-Conditioned Imitation Learning (GCIL) to learn general-purpose goal-specified policies from large uncurated datasets without rewards.
Our new goal-conditioned policy architecture "\textbf{BE}havior generation with \textbf{S}c\textbf{O}re-based Diffusion Policies" (BESO) leverages a generative, score-based diffusion model as its policy. 
BESO decouples the learning of the score model from the inference sampling process, and, hence 
allows for fast sampling strategies to generate goal-specified behavior in just 3 denoising steps, compared to 30+ steps of other diffusion-based policies.
Furthermore, BESO is highly expressive and can effectively capture multi-modality present in the solution space of the play data.
Unlike previous methods such as Latent Plans or C-Bet, BESO does not rely on complex hierarchical policies or additional clustering for effective goal-conditioned behavior learning. Finally, we show how BESO can even be used to learn a goal-independent policy from play-data using classifier-free guidance. To the best of our knowledge this is the first work that a) represents a behavior policy based on such a decoupled SDM  b) learns an SDM-based policy in the domain of GCIL and c) provides a way to simultaneously learn a goal-dependent and a goal-independent policy from play data.
We evaluate BESO through detailed simulation and show that it consistently outperforms several state-of-the-art goal-conditioned imitation learning methods on challenging benchmarks.
We additionally provide extensive ablation studies and experiments to demonstrate the effectiveness of our method for goal-conditioned behavior generation.
Demonstrations and Code are available at \url{https://intuitive-robots.github.io/beso-website}.
\end{abstract}

\IEEEpeerreviewmaketitle

\section{Introduction}

Goal-conditioned Behavior Learning aims to train versatile embodied agents, that can handle a wide range of daily tasks.
A common approach to tackle this challenge is Goal-conditioned Imitation Learning (GCIL).
GCIL only requires an offline dataset without additional rewards or expensive environment interactions for training.
However, GCIL typically requires a set of predefined tasks and a large number of labeled and segmented expert trajectories for each task, which can be costly and time-consuming.
Additionally, it does not generalize well to new scenes and different tasks.
Instead of teaching an agent a limited number of predefined goals, \textit{Learning from Play} (LfP) \cite{lynch2020learning} provides an effective way of collecting task-agnostic, teleoperated, uncurated, freeform datasets.
Such datasets consist of rich, meaningful, multimodal interactions with the environment that cover different areas of the state space.
Instead of manually labeling the trajectories, LfP pairs random sequences of each trajectory with one or more future states, i.e., the goal state,  of the respective trajectory.
Goal-conditioned policies distill useful, goal-oriented behavior from this collected play interaction data.
However, learning from play data remains an open challenge, partially due to the multimodal nature of the demonstrations, e.g., the same task can be solved in very different ways and different tasks can be solved in very similar ways.

Effective behavior learning from these datasets demands policies that maintain such multimodal solutions and that are expressive enough to remain close to the seen state-action distribution of the offline data for executing long-term horizon skills.
Most prior work tries to deal with this challenge, by combining generative models, such as Variational Autoencoders (VAEs) \cite{lynch2019play, mees2022calvin, pertsch2021accelerating} and Generative Pretrained Transformer (GPTs) \cite{cui2023from, shafiullah2022behavior}, with additional models and networks to explicitly encode multimodality or hierarchy.
However, these methods require supplementary networks or separation of skill execution and planning within their architecture, as the policy expression is not sufficient or cannot handle the multimodality of the observed behaviors.
Additionally, multiple learning objectives are typically required, e.g. for low- and high-level policies, which provides additional tuning challenges.

We propose a novel approach, \textbf{BE}havior Generation using \textbf{S}c\textbf{O}re-based Diffusion models (BESO), which excels in learning goal-conditioned policies solely from reward-free, offline datasets.
BESO uses Score-based Diffusion Models (SDMs) \cite{sohl2015deep, ho2020denoising, song2020score, karras2022elucidating}, a new class of generative models, that progressively diffuse data to noise through a forward Stochastic Differential Equation (SDE). 
By training a neural network, known as the score or denoising model, to approximate the score function, one can reverse the SDE to generate new samples from noise in an iterative sampling process.

We demonstrate several benefits of modeling the goal-conditioned action distribution using a score-based diffusion model.
First, we show, that the expressiveness of SDMs and their ability to capture multimodal distributions is key for effective conditioned behavior generation. 
On several challenging goal-conditioned benchmarks, including the conditioned Relay Kitchen and Block-Push environment \cite{cui2023from}, BESO consistently outperforms state-of-the-art methods such as C-BeT and Latent Motor Plans \cite{cui2023from, lynch2020learning}.
Second, by leveraging Classifier-Free Guidance Training of SDMs, BESO effectively learns two policies simultaneously: a goal-dependent policy and a goal-independent policy, which both can be used together or independently at test time.
Third, our model is easy and stable to train with a single training objective without additional rewards.
This contrasts with other state-of-the-art generative models, such as Implicit Behavior Cloning (IBC) \cite{florence2022implicit}, or hierarchical policies \cite{lynch2019play}.
Fourth, SDMs do not restrict the choice of the model architecture as in other generative models such as VAEs or energy-based models (EBMs) \cite{florence2022implicit}. 
Thus, we apply a novel Transformer architecture augmented with preconditioning to synthesize step-based actions given a sequence of observations and desired goal states.
Finally, BESO can diffuse new actions fast. 
While current diffusion-based policies \cite{pearce2023imitating} require $30+$ denoising steps for a single action prediction to achieve good results, our proposed approach, BESO, performs exceptionally well on challenging GCIL benchmarks, outperforming state-of-the-art goal-conditioned policies, while using only $3$ denoising steps.
We achieve this, by using recent advances in Score-based Diffusion Models, which separate the training and inference process \cite{karras2022elucidating} and applying novel numerical solvers designed for fast diffusion inference \cite{song2021denoising, lu2022dpm}.
Therefore, we systematically evaluate the essential components of SDMs for fast and effective step-based action generation.

To summarize our contributions:
\begin{itemize}
    \item BESO, a new policy representation based on score-based diffusion models for effective goal-conditioned behavior generation from uncurated play data 
    
    \item Use of Classifier-Free Guidance based Diffusion Policy to simultaneously learn a goal-dependent and goal-independent policy from play
    
    \item Systematic evaluation of key components for fast and efficient action generation using Score-based Diffusion policies combined with extensive experiments and ablation studies
\end{itemize}

\section{Related Work}

\begin{figure*}
    \centering
    \includegraphics[width=1\textwidth]{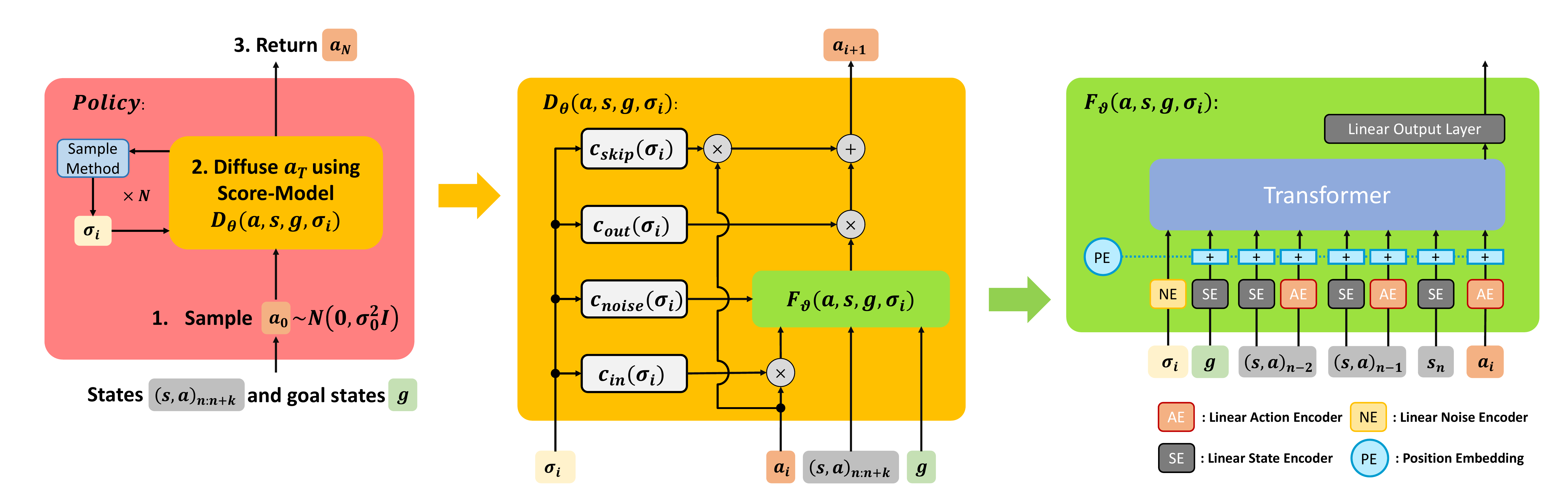}
    \caption{Overview of the action generation process of BESO with the used model architecture. \textbf{Left:} General Action Generation Process using the Diffusion Process to denoise the next action given the current observation $\state$ and desired goal state $\goal$ in $N$-steps. \textbf{Middle:} the high-level score-model with its pre-conditioning layers and skip connections. \textbf{Right:} the internal denoising score model, which uses a transformer architecture with causal masking to iteratively predict the denoised action given the sequence of prior observations, actions, and the goal sequence.}
    \label{fig: architecture overview}
\end{figure*}

\textbf{Diffusion Generative Models.}
Score-based generative models (SGMs) \cite{song2019generative, song2020improved} and Denoising Diffusion Probabilistic Models (DDPMs) \cite{sohl2015deep, ho2020denoising} are two different variants of score-based diffusion models (SDMs).
These models corrupt a data distribution with increasing Gaussian noise and use neural networks to learn to reverse this corruption to generate new data samples from noise. 
The two different models have been unified using the stochastic differential equation (SDE) framework \cite{song2020score}.
SDEs describe the diffusion process as a time-continuous process instead of using discrete noise levels. 
BESO follows the SDE formulation proposed by \citet{karras2022elucidating}.
To draw new samples from the diffusion models, they need to reverse the SDE discretized over $\Ndiff$ time steps.
The SDE contains a \textit{probability flow} ODE with the same marginal distributions, which allows for fast sampling \cite{song2020score}.
ODE solvers do not add noise during the inference process, which can reduce the number of function evaluations and accelerate sampling \cite{lu2022dpm}.
Sampling can be further accelerated using specialized numerical ODE solvers designed for diffusion inference \cite{ho2020denoising, karras2022elucidating, lu2022dpmsolver}.
SDMs achieved state-of-the-art results in various tasks including image generation \cite{karras2022elucidating}, text-based image synthesis \cite{dhariwal2021diffusion, rombach2022high} and human motion generation \cite{tevet2023human}.

\textbf{Goal-Conditioned Imitation Learning (GCIL).} It is a sub-domain of Imitation Learning \cite{osa2018algorithmic, argall2009survey}, where each demonstration is augmented with one or more goal-states that are indicative of the task that the demonstration was provided for.
The goal-state contains information that a learning method can leverage to disambiguate demonstrations.
Consequently, a goal-conditioned policy, i.e., a policy that includes the goal-state in its condition set, can use a given goal-state to adapt its behavior accordingly.
Similarly, goal-states have also extended the domain of reinforcement learning through Goal-Conditioned Reinforcement Learning (GCRL) \cite{eysenbach2022imitating, eysenbach2022contrastive, ma2022far, pertsch2021accelerating}, where the agent is not provided expert demonstrations but reward signals instead. Typically these reward signals are difficult to define, especially for complex tasks and environments, providing demonstrations is often a more natural option in such situations. 
Additionally, the policy rollouts required by GCRL are often expensive in real-world settings. 
Recent work investigated Goal Conditioned Offline Reinforcement Learning \cite{ma2022far, rosete-beas2022latent, pertsch2021accelerating, yang2021rethinking, mezghani2022learning}, which does not require these expensive rollouts during training.

\textbf{Learning from Play.}
The goal of \textit{Learning from Play} (LfP) \cite{lynch2020learning} is to learn goal-specified behavior from a diverse set of unlabeled state-action trajectories.
Classical imitation learning datasets typically consist of uni-modal, segmented expert trajectories in a narrow state-space. 
Play data, on the other hand, is characterized by unsegmented, multimodal trajectories. 
This makes learning meaningful behaviors more challenging, as the policies need the ability to deal with multiple ways of solving a task, distinguish between similar ways to solve different tasks, as well as the ability of long-horizon planning to reach goals far into the future. 
Prior work aimed to extract representations from play data for effective downstream policy learning \cite{9981307} or learned self-supervised representations of skills, referred to as latent plans, using Conditional Variational Autoencoders (CVAE) \cite{lynch2019play, lynch2020learning, mees2022calvin, mees2022hulc}. 
Transformer-based architectures were also researched as a policy class for task-agnostic behavior learning \cite{cui2023from, shafiullah2022behavior}.
Another body of work tries to improve LfP, by focusing on the data aspect and learning from object-centric interactions, instead of randomly sampled sequences \cite{belkhale2022plato}. 

\textbf{Generative Models in Policy Learning.}
Imitation Learning can be formulated as a state-occupancy matching problem, where the goal is to learn a policy that matches the state-occupancy distribution of expert demonstrations.
The unknown expert demonstration can now be approximated through modern generative model architectures.
One popular approach is the use of Generative Adversarial Networks (GANs) \cite{ho2016generative,fu2018learning}.
These methods consist of a generator policy that learns to imitate the observed behavior of the expert and a discriminator, which distinguishes between real and fake trajectories.
They require extensive rollouts during training, which is not feasible in our setting.
Other approaches use CVAEs \cite{Mandlekar-RSS-20, pertsch2021accelerating, lynch2019play, mees2022hulc, rosete-beas2022latent} to learn a latent embedding to represent the underlying skills. 
Recent work also applied Energy-based models as conditional policies for behavior cloning \cite{florence2022implicit}. 
Normalizing flows have also been proposed as a policy representation \cite{singh2020parrot}.

\textbf{Diffusion Generative Models in Robotics.} 
Most approaches that apply diffusion models in robotics applications focus on the discrete DDPM variant \cite{ho2020denoising}.
The DDPM Diffusion model has been used in Offline-RL to generate state-action or state-only trajectories using large U-net architectures \cite{janner2022diffuser, ajay2023is}.
DDPM has also been applied as a policy regularization method in a step-based Offline-RL setting in combination with a learned Q-function \cite{wang2023diffusion}.
Recently, score-based generative models have been leveraged to synthesize cost functions for grasp pose configurations \cite{urain2022se3dif}.
In addition, Conditional score-based generative models have been proposed to learn the reward function for inverse reinforcement learning \cite{kim2021imitation}.
The closest related work to BESO is Diffusion Policy \cite{chi2023diffusionpolicy} and Diffusion-BC \cite{pearce2023imitating}, which both propose the use of conditional, discrete DDPM as a new policy class for Behavior Cloning.
Diffusion-BC synthesizes new actions in $50$ stochastic sampling steps. 
To improve the performance, Diffusion-BC uses $X$-extra inference steps at the lowest noise level without additional noise.
However, this method results in even slower action generation.
BESO leverages the \textit{probability flow} ODE combined with fast, deterministic samplers and optimized noise levels.
Hence, BESO requires significantly fewer function evaluations in every action prediction.

\section{Problem Formulation and Method}
In this section, we describe our approach to goal-conditioned behavior generation using Score-based diffusion models. 

\subsection{Problem Formulation}

The Goal of GCIL is to learn a general-purpose goal-conditioned policy from uncurated play data. 
Given a set of unstructured, task-agnostic trajectories, $\Tset = \left\{\traj_\itraj|\traj_\itraj=((\state_\itime^\itraj,\act_\itime^\itraj))_{\itime=1}^{\Ntime_\itraj}\right\}$, each trajectory can be split into a set of tuples containing sub-trajectory sequences and goal-states $\Dset_\itraj=\left\{(\seq,\goal) | \seq=(\state_\itime,\act_\itime)_{\itime=i}^{i_\dag},\goal=(\state_\itime)_{\itime=j}^{j_\dag},(\state_\itime,\act_\itime)\in\traj_\itraj\right\}$, with $i \le i_\dag < j \le j_\dag$ denoting start and end steps of the sequence and goal-state respectively.
As this formulation makes clear, the goal-state has to be one or more states of the same trajectory as the sequence and has to begin at some step after the respective sequence has ended.
The set $\Dset_\itraj$ can contain overlapping sequences and the final play dataset is given as $\Dset = \bigcup_{\itraj=1}^\Ntraj \Dset_\itraj$. For simplicity, the indices of $\seq_\itraj$ and $\goal_\itraj$ simply indicate that the sequence and goal state belong together and the indices in $(\state_\itime,\act_\itime) \in \seq$ refer to the relative time step in the sequence.
The state-action pairs in the sequence $\seq_\itraj$ leading to the goal state $\goal_\itraj$ are now treated as the optimal behavior to reach $\goal_\itraj$\cite{lynch2019play, mees2022calvin}.
Goal-conditioned policies try to maximize the log-likelihood objective over the play dataset
\begin{align}
    \mathcal{L}_{\text{play}} = \mathbb{E}_{(\seq, \goal) \in \Dset} \left[ \sum_{(\state,\act) \in \seq} \log \p{\pi_\theta}{\act | \state, \goal}  \right].
\end{align}
Because of the multi-modal nature of the demonstrations, i.e. several trajectories leading to the same goal state, solving this objective successfully requires a policy that is capable of encoding such a multi-modal behavior.

\subsection{Score-based Diffusion Policies}

We now aim to learn the policy distribution $\p{\pi_\Dset}{\act | \state, \goal}$ underlying the play dataset $\Dset$ and, hence, the given demonstrations.
We do so by defining a continuous diffusion process, which maps samples from our play dataset by gradually adding Gaussian noise to the intermediate distributions $p_\idiff, \idiff\in [0,\Ndiff]$ with initial distribution $p_0 = \pi_\Dset$ and final distribution $p_T$.

The continuous diffusion process can be described using a stochastic-differential equation (SDE) \cite{song2020score}. 
In this work, we define the SDE analogously to a recently introduced formulation \cite{karras2022elucidating}:
\begin{equation}
\label{eq: conditional Karras Song SDE}
\begin{split}
d \mathbf{a} =  \big( \beta_t \sigma_t - \dot{\sigma}_t  \big) \sigma_t \nabla_a \log p_t(\act | \state, \goal) dt + \sqrt{2 \beta_t} \sigma_t d\omega_t,
 \end{split}
\end{equation}
where $\nabla_{\act} \log p_t(\act | \state, \goal)$ refers to the score-function, $\omega_t$ is the Standard Wiener process, which can be understood as infinitesimal Gaussian noise.
The noise scheduler is denoted by $\sigma_t$, and $\beta(t)$ describes the relative rate at which the current noise is replaced by new noise. 
In our approach, we adopt $\sigma_t(t) = t$, a method proven effective in image generation \cite{karras2022elucidating}.
At every timestep $t$ and related noise level there exists a corresponding marginal distribution $p_t(\act| \state, \goal)$, which is the result of injecting Gaussian noise to samples from $p_{\text{play}}$.
 This can be expressed as $p_{t}(\act_t| \act) = \mathcal{N}(\act, \sigma_t^2 \mathbf{I})$.
The final action distribution of the diffusion process is a known tractable prior distribution $\act_T = p_T$. 
An unstructured Gaussian distribution $p_T = \mathcal{N}(\mathbf{0}, \sigma_T^2 \mathbf{I})$ is chosen without any information about the play data distribution.

In the case of BESO we are particularly interested in the \textit{Probability Flow} Ordinary Differential Equation (ODE) within the SDE \cite{chen2018neural}.
This ODE shares the same marginal distributions $p_t(\act |\state, \goal)$ as the SDE at every timestep, but without the additional random noise injections. 
By setting $\beta(t) = 0 $, we recover the \textit{Probability Flow} ODE from Eq. \eqref{eq: conditional Karras Song SDE}:
\begin{equation}
\label{eq: karras ODE}
d \act = - \dot{\sigma}_t \sigma_t \nabla_{\act} \log p_t(\act| \state, \goal)dt
\end{equation}
The negative score-function $-\nabla_{\act} \log p_t(\act| \state, \goal)$ specifies the vector field of the current marginal distribution $p_t(\act | \state, \goal)$. This vector field points towards regions of low data density and is scaled with the product of the current noise level $\dot{\sigma}_t$ and the change of it $\dot{\sigma}_t$.

\begin{algorithm}
\caption{BESO Training}
    \label{alg: beso training}
    \begin{algorithmic}[1]
    \State \textbf{Require:} Play Dataset $\mathcal{L}_{\text{play}}$, Sequence Size $c_o$, Goal Sequence Size $c_g$
    \State \textbf{Require:} Score Model $\mathbf{D}_{\theta}(\act, \state, \goal, \sigma_t)$
    \State \textbf{Require:} Noise Distribution $\sigma \sim \text{LogLogistic}(\alpha, \beta)$
    \For{$i \in \{0, ...,N_{\text{train steps}} \}$ }
        \State Sample $(\seq, \goal) \sim \mathcal{L}_{\text{play}}$ 
        \State Sample $\boldsymbol{\epsilon} \sim \mathcal{N}(\sigma_{\text{mean}}, \sigma_{\text{std}}^2 \mathbf{I})$
        
        \State $ \mathcal{L}_{D_{\theta}} \gets
     \mathbb{E}_{\mathbf{\sigma}, \act, \boldsymbol{\epsilon}} \big[  \alpha (\sigma_t) \| D_{\theta}(\act + \boldsymbol{\epsilon}, \state, \goal, \sigma_t)  - \act  \|_2^2 \big] $
    \EndFor
    \end{algorithmic}
\end{algorithm}

\subsection{Diffusion Training}
In order to generate new samples by numerically approximating the reverse ODE, we require an accurate estimate of the score function $\nabla_{\act} \log p_t(\act| \state, \goal)$ for all marginal distributions $p_t$ in our diffusion process.
To achieve this, we use a neural network $D_{\theta}(\act, \state, \goal, \sigma_t)$ that matches the score for all marginal distributions $p_t(\act |\state, \goal)$.
\begin{equation}
\label{eq: score function approx}
\nabla_{\act} \log p_{t}(\act | \state, \goal) =  \left( D_{\theta}(\act, \state, \goal,  \sigma_t) - \act \right)/ \sigma_t^2.
\end{equation}
The neural network is trained using the denoising score matching objective \cite{6795935, song2019generative}, where we add Gaussian noise to the actions and minimize the difference between the network's output and the original actions:
\begin{equation}
\label{eq: denoising score matching loss}
\begin{split}
     \mathcal{L}_{D_{\theta}} =  
     \mathbb{E}_{\mathbf{\sigma_t}, \act, \boldsymbol{\epsilon}} \left[  \alpha (\sigma_t) \| D_{\theta}(\act + \boldsymbol{\epsilon}, \state, \goal, \sigma_t)  - \act  \|_2^2 \right],
     \end{split}
\end{equation}
where $\act$ is an action sample, and $\boldsymbol{\epsilon} \sim \mathcal{N}(\mathbf{0}, \sigma_t^2 \mathbf{I})$ represents the Gaussian noise. 
The losses at individual noise levels are weighted according to $\alpha(\sigma_t)$, and the current $\sigma_t$ is sampled from the noise training distribution $p_{\text{train}}$.
We use a truncated log-logistic distribution with location parameter $\alpha$ and scale parameter $\beta$: $p_{\text{train}} \sim \text{LogLogistic}(\alpha, \beta)$ in the range of $\{ \sigma_{\text{min}},\sigma_{\text{max}} \}$.
The training process is summarized in Alg. \ref{alg: beso training}.
This allows us to effectively learn the noise-conditioned score function for our diffusion process and generate samples from the conditional density, $p_t(\act|\state, \goal)$, using the Probability Flow ODE. 

\subsection{Efficient Action Generation using Deterministic Samplers}
\label{sec: ode generation}
New actions are generated by our policy by sampling from the prior distribution $\act_T \sim \mathcal{N}(\mathbf{0}, \sigma_T^2 \mathbf{I})$ and numerically simulating the reverse ODE or SDE by substituting the score-function with our learned model in Eq. \eqref{eq: karras ODE}. 
The process begins by selecting a random sample from our prior distribution,  $\act_T \sim \mathcal{N}(\mathbf{0}, \sigma_T^2 \mathbf{I})$, and then iteratively denoise this sample.
Utilizing a random sample as a starting point enables the creation of diverse and multimodal actions, even when the underlying ODE is deterministic.
The ODE can be solved numerically, by discretizing the differential equation starting from $T$ to $0$. 
During the action prediction, we iteratively denoise the sample at $N$-discrete noise levels. 
BESO employs the DDIM solver, as described in detail in Alg. \ref{alg: DPM-1} \cite{lu2022dpmsolver, song2021denoising}, for fast, deterministic sampling. 
The solver is a first-order deterministic sampler that is based on an exponential integrator method.
A detailed comparison of state-of-the-art diffusion samplers is provided in Sec. \ref{sec: Sampler Ablation studies} of the Appendix, which concludes, that DDIM has the best overall performance.
An additional evaluation on the influence of noise concludes that ODE solvers are competitive with SDE variants for action prediction tasks.
Our ablation studies in Sec. \ref{sec: Sampler Ablation studies} suggest that only three denoising steps are necessary for BESO to generate actions with high accuracy.
Increasing the number of inference steps further only marginally enhances the performance, while significantly slowing down the sampling process.
Thus, we found that 3 steps strike the best balance between computational efficiency and performance.
For inference, we can adapt the range of noise and the distribution of discrete timesteps. 
Based on empirical evaluations, we decide to use exponential time steps with a noise range of $\sigma \in \{ 0.005, 1 \}$ for most applications. 
\begin{algorithm}
    \caption{Action Generation Process using DDIM based Sampler (DPM-1) adapted for BESO \cite{lu2022dpm, song2021denoising} }
    \begin{algorithmic}[1] 
    \State \textbf{Require:} Current state $\state$, goal $\goal$
    \State \textbf{Require:} Score-Denoising Model $D_{\theta}(\act, \state,\goal, \sigma)$
    \State \textbf{Require:} Discrete time steps $t_{i \in \{ 0,.., N\}}$
    \State \textbf{Require:} Noise scheduler $\sigma_i = t_i$
    \State \textbf{Require:} $f_{\beta}(t) =  - \log (t)  $, $f_{t}(\beta) =   \log ( - \beta)  $
    \State Draw sample $\act_0 \sim \mathcal{N}( \mathbf{0}, \sigma_{0}^2\mathbf{I})$ 
        \For{$i \in \{0, ...,N-1 \}$ }
            \State{ $\mathbf{d}_{i} \gets \big( \act_i - D_{\theta}(\act_i, \state,\goal, \sigma_{i}) \big) / \sigma_{i}$} 
            \State $\beta_{t_i}, \beta_{t_{i+1}} \gets f_{ \beta}(t_i),  f_{ \beta}(t_{i+1})$
            \State $h_i \gets \beta_{t_i} - \beta_{t_{i-1}}$
            \State $\act_{i+1} \gets (\frac{t_{i+1}}{t_{i}}) \act_{i} - \big(e^{(-h_{i})} - 1 \big)\mathbf{d}_{i}$
        \EndFor
\State \textbf{return} $\mathbf{a}_N$
    \end{algorithmic}
    \label{alg: DPM-1}
\end{algorithm}

\section{Goal-Guided Score-based Diffusion Policies}

In this section, we introduce two variants of BESO optimized for synthesizing actions for goal-conditioned behavior. 

\textbf{Conditioned Policy (C-BESO).}
We define a goal-conditioned diffusion policy, $\gpol$, by directly learning the goal-and-state-conditioned distribution with our score-based generative model.
In contrast to standard goal-conditioned behavior cloning, our diffusion policy allows us to capture multiple solutions present in the play data while still being expressive enough to solve long-term goals.

\textbf{Goal-Classifier-Free Guided Policy (CFG-BESO).}
We additionally combine BESO with a popular conditioning method for diffusion models, Classifier-Free Guidance (CFG) \cite{ho2021classifier}.
We train a goal-conditioned diffusion policy $\gpol$ by applying a dropout rate of $0.1$ to the goal $\goal$, which also trains an implicit goal-independent policy $\pol$ within our goal-conditioned model. 
The generation process uses a combined gradient for the denoising process
\begin{equation}
\label{eq: classifier-free guidance}
\begin{split}
    \nabla_{a} \log p_{t,\lambda}(\act|\state, \goal) =\\ \lambda \nabla_a \log p_t(\act | \state, \goal) + (1 - \lambda) \nabla_a \log p_t(\act| \state ),
\end{split}
\end{equation}  
where the guidance factor $\lambda$ balances the influence of the goal-conditioned and goal-independent gradient.
In diffusion literature, $\lambda$ commonly ranges from $2$ to $7.5$, to guide the diffusion model towards goal-conditional distribution $\gpol$.
CFG has demonstrated significant performance improvements compared to other conditioning methods \cite{ho2021classifier, lu2022dpm, pmlr-v162-nichol22a}. 
Even though CFG has also been successfully applied for generating state-only trajectories in Offline-RL \cite{ajay2023is}, recent work on behavioral cloning suggests that CFG performs significantly worse than simpler conditioning methods \cite{pearce2023imitating} for step-based action generation.
We provide a detailed analysis of CFG for goal-guided action generation in our experiment section.
\begin{table*}
        \centering
        \scalebox{0.9}{
        \begin{tabular}{ll|cccccccc}  
        \toprule
       &    &  GCBC & C-IBC  & LMP & RIL  & C-BeT & CX-Diff &  C-BESO  & CFG-BESO   \\
             \midrule
    \multirow{2}{*}{Block-Push} & Reward   & 0.13 ($\pm$ 0.04)  &  0.46 ($\pm$ 0.06) & 0.04 ($\pm$ 0.03) & 0.06 ($\pm$ 0.01) & 0.91 ($\pm$ 0.03) & 0.93 ($\pm$ 0.03) &  \textit{0.96} ($\pm$ 0.02)  & \textit{\textbf{0.97}} ($\pm$ 0.02) \\   
                                & Result      & 0.13 ($\pm$ 0.04)  & 0.29 ($\pm$ 0.10) & 0.04 ($\pm$ 0.03) & 0.02 ($\pm$ 0.01) & 0.87 ($\pm$ 0.07)  & 0.90 ($\pm$ 0.04) &   \textit{\textbf{0.93}} ($\pm$ 0.02)  & 0.88 ($\pm$ 0.04) \\   
    \midrule
    \multirow{2}{*}{Relay-Kitchen} & Reward   & 2.65 ($\pm$ 0.25) & 0.50 ($\pm$ 0.09) & 1.45 ($\pm$ 0.22) & 0.31 ($\pm$ 0.15) &  2.73 ($\pm$ 0.28)  & 3.64 ($\pm$ 0.14) & \textit{\textbf{3.98}} ($\pm$ 0.07)  & \textit{\textbf{3.98}} ($\pm$ 0.07)  \\    
                                & Result      & 2.57 ($\pm$ 0.26) &  0.45 ($\pm$ 0.08) & 1.41 ($\pm$ 0.22) & 0.23 ($\pm$ 0.11) & 2.69 ($\pm$ 0.28) & 3.35 ($\pm$ 0.15) & \textit{\textbf{3.75}} ($\pm$ 0.08)  & \textit{3.47} ($\pm$ 0.08) \\
            \bottomrule
        \end{tabular}        }
        \caption{Mean and std on the conditioned block-push and kitchen environment, over 10 seeds with 100 runs each. C-BESO and CFG-BESO consistently outperformed all baselines, despite only using 3 inference steps. CX-Diff with 3 inference steps achieves a result of 2.74(±0.26) on the relay-kitchen. Both variants of BESO show a low deviation across seeds, indicating their robustness.
        }
        \label{tab: GC Simulation Benchmark}
\end{table*}
\subsection{Model Architecture}
One of the main challenges of training the score-based diffusion model is the big range of noise levels $\sigma_t \in \{ 0.001, 40 \}$
To address this challenge, we use an improved architecture \cite{karras2022elucidating} including additional skip-connections and two pre-conditioning layers, which are conditioned on the current noise level $\sigma_t$ 
\begin{equation}
\begin{split}
    D_{\theta}(\act| \state, \goal, \sigma_t) =\\ c_{\text{skip}}(\sigma_t) \act + c_{\text{out}}(\sigma_t)F_{\theta}(c_{\text{in}}(\sigma_t)\act, \state, \goal, c_{\text{noise}}(\sigma_t)),
\end{split}
\end{equation}
The conditioning functions are described in detail in Section \ref{sec: BESO Hyperparameters} of the Appendix and visualized in Figure \ref{fig: architecture overview}.

These additional skip connections help the score model to scale the output to a wide range of noise levels $\sigma_t$, either by estimating the denoised sample $\act_{t-1}$, directly predicting the noise $\mathbf{\epsilon}$ or something in between these two.  
Our proposed approach, BESO, integrates a Transformer-based architecture with causal masking as the inner model $F_{\theta}(\act, \state,\goal, \sigma_t)$.
This enables our model to learn temporal relations between observations and actions, thereby improving its overall performance
A detailed overview of our proposed architecture is shown in Figure \ref{fig: architecture overview}.
Three linear embedding layers encode the states $\state_n$, noise $\sigma_t$ and the noisy actions $\act_n$ into a linear representation of the same dimension, $l_{\state}(\state), l_{\act}(\act),l_{\sigma}(\sigma)$. 
In addition, the position embedding information is added on the linear representations. 
The noise embedding is concatenated with the desired future states and all state-noise-action pairs in a large sequence for the model. 
During training, the denoised actions are inferred for all timesteps in the input series, yet only the last predicted action is utilized for inference.
To take advantage of the causal masking in the transformer, we concatenate the goal-sequence before the current observation sequence \cite{cui2023from}, allowing for a sequence of goal-states.

\section{Evaluation}
    
\begin{figure}
    \centering
    \includegraphics[width=0.488\textwidth]{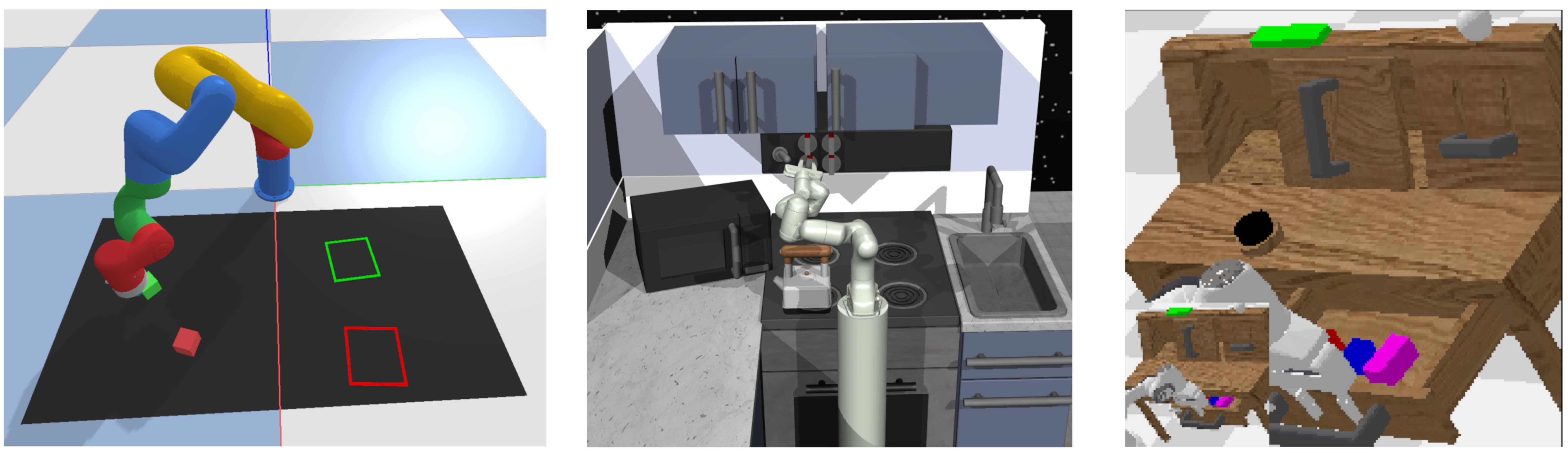}
    \caption{Simulation environments for testing the performance of BESO: 
    Multi-Modal Block-push (left); Relay Kitchen (middle); CALVIN (right)
    }
    \label{fig: Experiments Overview}
\end{figure}
The objective of our experiments was to answer the following key questions: 
\textbf{I}) Is BESO competitive on goal-conditioned environments against state-of-the-art baselines? \textbf{II}) What are the key components to enable fast sampling of Diffusion policies with good performance? \textbf{III}) Does Classifier-Free Guidance work for goal-conditional behavior synthesis?
To answer these questions, we evaluated BESO on several challenging simulation benchmarks.
First, we compared the performance of BESO against other state-of-the-art methods. Afterward, we examined BESO's components with respect to their contribution to the performance.

\subsection{Baselines}
We compare BESO against several state-of-the-art methods:
\begin{itemize}
    
\item  \textbf{Goal-conditioned Behavior Cloning (GCBC)} learns a unimodal policy encoded as a simple multi-layer perceptron (MLP) with an trained with an MSE loss \cite{lynch2020learning}.

\item \textbf{Relay Imitation Learning (RIL)} is a hierarchical policy, that learns a high-level sub-goal generator, which is used to condition a low-level MLP policy \cite{lynch2019play}. 

\item \textbf{Latent Motor Plans (LMP)} 
    is a hierarchical goal-conditioned policy, which consists of a seq2seq CVAE and an action decoder policy \cite{lynch2020learning}. 
    We use an adapted KL-weighting term and a transformer encoder, which has been shown to improve the performance of LMP \cite{mees2022hulc}. 

\item \textbf{Conditional Implicit Behavior Cloning (C-IBC)} uses an energy-based model as an implicit policy \cite{florence2022implicit}. We use a goal-conditioned extension of IBC to study the importance of the selected generative model architecture.

\item \textbf{Conditional-Behavior Transformer (C-BeT)} is a GPT-like transformer-based policy, that predicts discrete action labels together with a continuous offset vector to learn multimodal behavior \cite{shafiullah2022behavior, cui2023from}. The action labels are determined a priori via K-means clustering.

\item \textbf{Diffusion-X (CX-Diff)} \cite{pearce2023imitating}
    is a DDPM \cite{ho2020denoising} based policy with improved inference. It uses stochastic sampling and additional $X$-extra inference steps at the lowest noise level to synthesize actions in $50$+$X$ steps. While performing only slightly worse than the closely related KDE-Diff \cite{pearce2023imitating} it has a significantly lower computational cost.\end{itemize}

To ensure a fair evaluation of all methods we kept the general hyperparameters, e.g., layer size and number, as consistent as possible while tuning the method-specific hyperparameters. A detailed summary of the baseline architectures and hyperparameters is provided in Sec. \ref{sec: baseline implementation} of the Appendix.
Additionally, we evaluated all models on the kitchen and block-push task with 10 seeds and 100 rollouts each.
Given the high computational costs and time of training models for CALVIN, we restricted the tested methods to 3 seeds and limited the number of baselines.

\subsection{Simulation Experiments}

\begin{figure*}
    \centering
    \includegraphics[]{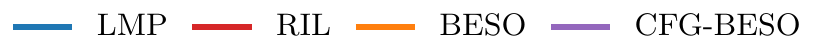}
    \includegraphics[width=1\textwidth]{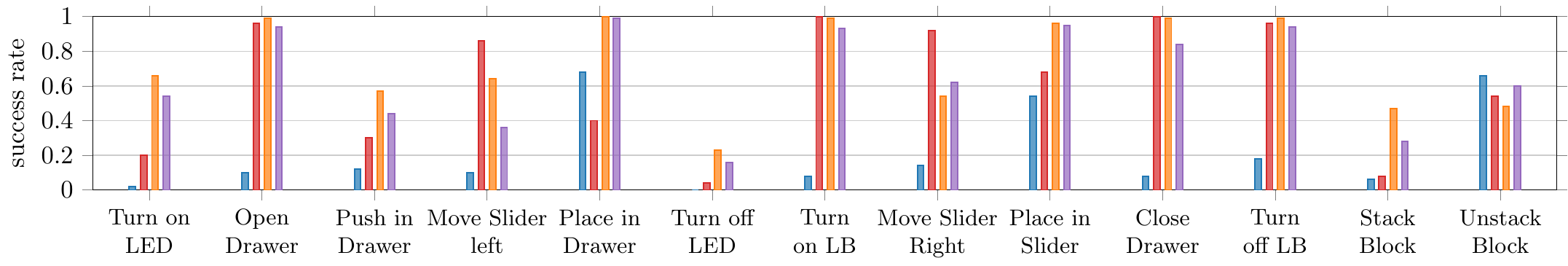}
    \caption{The average Success rate of all tested models on executing single hard tasks in the CALVIN environment conditioned on a single goal image, that does not contain the end-effector of the robot near the required task. }
    \label{fig: CALVIN Hard tasks}
\end{figure*}
We evaluated BESO against the baselines on three simulation benchmarks, shown in Figure \ref{fig: Experiments Overview}:
\begin{itemize}
    \item \textbf{CALVIN Benchmark \cite{mees2022calvin}:} We used the LfP benchmark, with a dataset consisting of 6 hours of unstructured play data. 
    We restricted all methods to using a single static RBG image as observation input and predicting relative Cartesian actions as output \cite{rosete-beas2022latent}.
    We evaluated the methods on single tasks and 2 tasks in a row from a single goal image, both variants were conditioned on goal-images outside the training distribution, that did not contain the end-effector in the correct position.
    \item \textbf{Block-Push Environment \cite{florence2022implicit}:} We used the adapted goal-conditioned variant \cite{cui2023from}.
    The Block-Push Environment consists of an XARm robot that must push two blocks, a red and a green one, into a red and green squared target area.
    The dataset consists of 1000 demonstrations collected by a deterministic controller with 4 possible goal configurations.
    The methods got 0.5 credit for every block pushed into one of the targets with a maximum score of 1.0.
    \item \textbf{Relay Kitchen Environment \cite{lynch2019play}:}
    A multi-task kitchen environment with objects such as a kettle, door, and lights that the agent can interact with.
    The data consists of 566 human-collected trajectories with sequences of 4 executed skills.
    We used the same experiment settings as described in \cite{cui2023from} to allow for fair comparisons.
    The models were evaluated using a pre-defined goal state, that consisted of 4 tasks for each rollout.
    Each correctly completed task gives 1 credit with a maximum of 4.
\end{itemize}
The methods were evaluated on two metrics:
\textbf{result} evaluates how many of the desired goals of each rollout are achieved, while \textbf{reward} measures the overall performance by giving credit for reaching any goal defined in the environment.

\subsection{Simulation Results}
We compared BESO to the baselines on the Relay-Kitchen and Block-Push environments. 
The results are summarized in Table \ref{tab: GC Simulation Benchmark}.
As shown in the table, BESO consistently outperformed the competitors on both tasks across 10 seeds. The low variance of BESO, additionally, indicates the robustness of our approach.
Among the baselines, Diffusion-X and C-BeT perform well on the kitchen task and block-push environment, respectively.
The diffusion policies excelled, outperforming all other baselines on the kitchen and the block-push task, whereas C-BeT demonstrated comparable performance on the block-push environment.
Considering that BESO only used $3$ denoising steps on both environments, compared to the $50(20)+8$ steps of CX-Diff, makes BESO's performance even more impressive. 
By contrast, CX-Diff, when limited to $3$ denoising steps, only managed an average result of $2.74 (\pm 0.26)$ in the kitchen environment.
This highlights the advantage of BESO's architecture combined with improved noise scheduling and sampler to achieve good results with only 3 denoising steps.
On a modern desktop PC, BESO requires around $0.012$ seconds to predict an action, while the CX-Diffusion model needs an average of $0.15$ seconds. 
This makes BESO over $10$ times faster.

\begin{figure}
    \centering
    \includegraphics[width=0.5\textwidth]{plots/rss_23_calvin_legend.pdf}
    \includegraphics[width=0.5\textwidth]{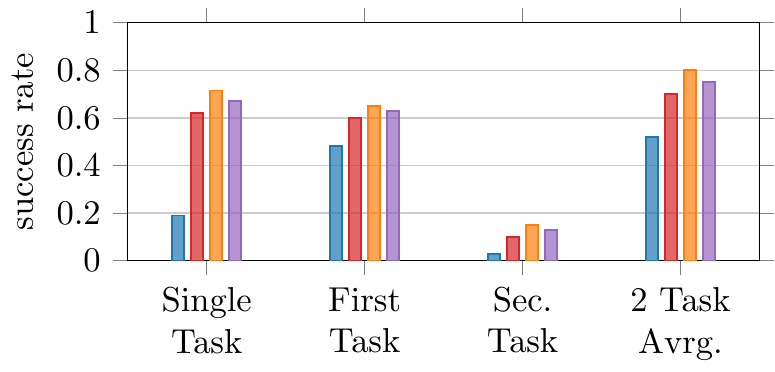}
    \caption{The average performance of goal-conditioned policy on the CALVIN environment. The first column shows the average success rate of 13 individual tasks. The other three columns show the average success rate of all models conditioned on a single goal image with 2 tasks.}
    \label{fig: calvin results}
\end{figure}

In a more challenging simulation environment, the CALVIN environment, BESO demonstrated its ability to generalize to unseen goal states by achieving the best overall performance on 13 difficult single tasks.
Each task was conditioned on a single goal image unseen during training, where the end-effector is not located near the corresponding task.
This posed a significant challenge, as the models have to infer changes in the environmental state and perform the necessary tasks without relying on the position of the end-effector in the image for guidance.
The results of this experiment are summarized in Figure \ref{fig: calvin results} and the individual success rates of the tasks are summarized in Figure \ref{fig: CALVIN Hard tasks}. 
As shown, BESO achieves the best overall performance on individual hard tasks, demonstrating its ability to also generalize to unseen goal-states.
RIL is the second-best model and has a slightly better average performance on 2 tasks.

Additionally, the models were evaluated on solving two tasks with a single goal image.
Similar to the first task, the end-effector was located at a different position away from both tasks.
 In this instance, BESO and its Classifier-Free Guidance (CFG) variant once again outperformed other models, though the CFG variant registered a slightly lower performance.
The results illustrate that BESO can effectively learn meaningful behavior to solve downstream short-term and long-term goals by learning from random windows of play trajectories. 
This further supports the conclusion that BESO's ability to learn multimodal and expressive action distributions is key for effective learning from play.
In addition, this experiment showcases BESO's proficiency in effectively from visual data. 
Overall, our results indicate that BESO is competitive against state-of-the-art baselines and capable of effectively learning from play data, making it a promising approach for goal-conditioned behavior learning. Hence, we can answer Question \textbf{I}) in the affirmative.

\subsection{BESO design choices}
We answer Question \textbf{II} by evaluating different components of BESO to study their contribution to the overall performance.

\textbf{Conditioning Method.}
First, we evaluated different methods to condition the behavior generation on the desired goal state.
We tested the FiLM-conditioning \cite{perez2018film} and the sequential conditioning method used in C-BeT \cite{cui2023from}.
FiLM requires additional MLP models, which input the goal and scale the latent representations inside the transformer layers.
The sequential conditioning method simply includes desired goal-states at the beginning of our sequence as depicted in the model overview of Figure \ref{fig: architecture overview}.
We tested both conditioning variants using the same transformer score model and evaluated it on the block-push and kitchen environment on 10 seeds.
FiLM conditioning resulted in a performance drop compared to the sequential conditioning method from an average result of $0.93$ to $0.91$ and $3.76$ to $3.4$ on the block-push and kitchen environment respectively. 
Moreover, the FiLM method increase the overall model capacity.
Hence, BESO uses the sequential conditioning method.

\vspace{\baselineskip}
\textbf{Sampling Algorithm.}
BESO generates actions by numerically approximating the reverse ODE with its learned score-model starting from a sample generated from our Gaussian prior distribution $p_{T}$. 
We investigated several numerical sampling algorithms used in diffusion research, such as DDIM \cite{song2021denoising}, DPM \cite{lu2022dpm}, DPM++ \cite{lu2022dpmsolver}, and Heun \cite{karras2022elucidating}, to assess their contribution to BESO's performance.
The samplers were evaluated on the block-push and kitchen environments with different number of denoising steps. 
The results show that the performance gap between the individual samplers is small, with DDIM achieving the best overall performance.
Surprisingly, the second-order Heun solver has the worst average performance.
Detailed results of this experiment are summarized in Table \ref{tab: deterministic kitchen sampler study} and Table \ref{tab: deterministic push sampler study} in the Appendix.
Overall BESO is robust to the number of sampling steps and chosen sampler type, maintaining a similar performance from $3$ to $50$ inference steps.

\begin{table}
    \centering
    \begin{tabular}{ll|cc}
    & & Deterministic  & Stochastic \\
    \midrule
      \multirow{2}{*}{\makecell{Block-\\Push}}    & Reward & 0.97 ($\pm$ 0.02) & 0.97 ($\pm$ 0.02) \\
                                     & Result & 0.93 ($\pm$ 0.02) & 0.92 ($\pm$ 0.03) \\
    \midrule
      \multirow{2}{*}{\makecell{Relay-\\Kitchen}} & Reward & 3.95 ($\pm$ 0.10) & 4.03 ($\pm$ 0.07) \\
                                     & Result & 3.73 ($\pm$ 0.11) & 3.80 ($\pm$ 0.08)  \\
    \midrule
    \multirow{2}{*}{\makecell{CALVIN}} & Hard Tasks & 0.71 ($\pm$ 0.01) & 0.68 ($\pm$ 0.03) \\
                                     & 2 Tasks & 0.79 ($\pm$ 0.03) & 0.79 ($\pm$ 0.02)  \\
    \bottomrule
    \end{tabular}
    \caption{Evaluation of the Influence of Noise Injection for Goal-conditional Behavior Generation averaged over 2 samplers with and without random noise injection. 
    }
    \label{tab: noise comparison}
\end{table}

\newpage
\textbf{Stochastic vs. Deterministic Sampling.}
Current diffusion literature supports the assumption that stochastic samplers have a better overall performance compared to deterministic samplers \cite{karras2022elucidating, song2020score}.
We tested this assumption with respect to step-based action generation.
We evaluated the same models with 2 sampling algorithms DPM++(2S) and the Euler sampler \cite{lu2022dpm, karras2022elucidating}, each with and without noise injection.
The noise scheduling was performed via the ancestral sampling strategy, as used in the DPPM variant \cite{ho2020denoising, song2020score} and described in Alg. \ref{alg: Ancestral Sampling}.
Experiments were again conducted in all environments.
As shown in Table \ref{tab: noise comparison}, the results suggest that the addition of noise does not offer a significant benefit to the action generation of step-based diffusion policies.
Stochastic samplers only increase the average performance in the kitchen environment.
The discrepancy compared to common diffusion applications such as image synthesis \cite{dhariwal2021diffusion} could be rooted in high-dimensional image spaces, making the generation process more difficult and requiring more steps for good results. In these high-dimensional spaces, errors are more likely to occur and accumulate over time. Adding noise during the inference process helps the model to correct errors of the gradient approximation, resulting in a better overall performance \cite{karras2022elucidating}.
In contrast, step-based action-distributions are significantly lower dimensional than the high-dimensional latent spaces of image generation, hence, the addition of noise does not appear to benefit the average performance of step-based policies, as supported by our experimental results.

\vspace{\baselineskip}
\textbf{Classifier-Free-Guidance (CFG).}
\begin{figure}
    \centering
    \includegraphics[width=0.45\textwidth]{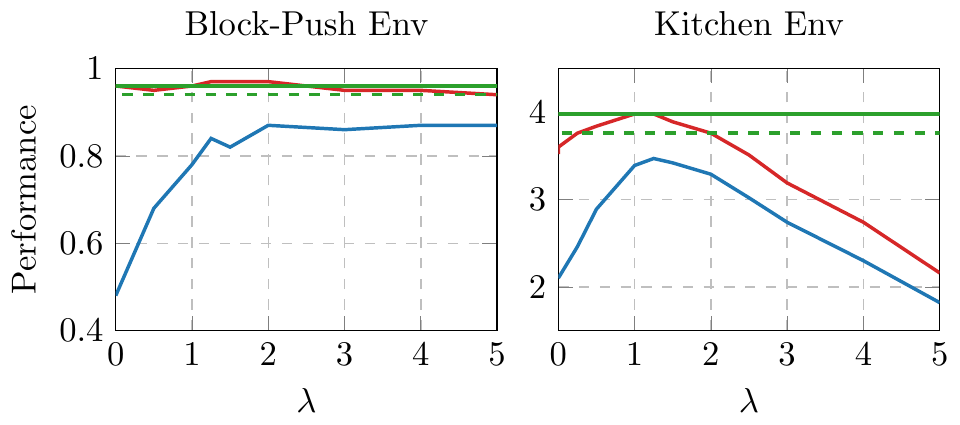}
    \includegraphics[width=0.4\textwidth]{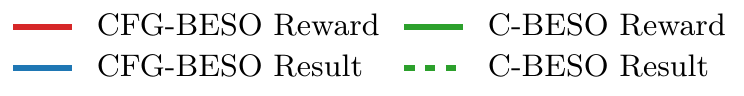}
    \caption{Comparison of CFG Method for Goal-conditioned Behavior Learning from Play Data. 
    For CFG-BESO we evaluate the 10 seeds on 100 rollouts each with different $\lambda$ values. 
    The CFG variant of BESO has a slightly worse average result in both environments with similar rewards.
    When using $\lambda=0$, we can recover an unconditional policy, that performs random rollouts with high rewards and low results.
    }
    \label{fig: cfg_results}
\end{figure}
Finally, we investigate Question \textbf{III} by evaluating the effect of Classifier Free Guidance (CFG) for step-based action generation with goal-conditioned policies.
The results of this experiment, reported in Figure \ref{fig: cfg_results}, indicate that CFG is an effective method for goal-conditioning in a step-based setting.
The average result for the block-push and kitchen tasks is slightly worse than the standard goal-conditioned variant, while the average reward is equal.
CFG-BESO is also able to learn effectively in the image-based CALVIN environment and achieves similar performance to the standard goal-conditioned variant. 
The performance of the CFG-model with $\lambda=0$ demonstrates, that CFG-BESO is capable of learning a well-performing, unconditional policy $\pi \left(\act| \state\right)$. 
The low average result in Figure \ref{fig: cfg_results} shows that the policy ignores the goal-state and aims to achieve a high reward solely based on the current state.
This gives CFG-BESO a unique advantage over common play-based policies. 
However, CFG has a trade-off: it slightly lowers the average result for more diverse rollouts.
Empirical evaluations suggest the best $\lambda$ value is $1.25$ for most tested environments. 
Experiments with higher values resulted in a lower average performance in environments with high-dimensional action spaces, indicating instability in the action generation.
We hypothesize that the guidance provided by the goal-conditioning is only crucial in certain steps during the rollouts, specifically when the policy is deciding which task to solve.

\section{Conclusion} 
\label{sec:conclusion}

We introduced BESO, a new policy representation for goal-conditioned behavior generation that uses score-based diffusion models. 
We leveraged the expressiveness and multimodal properties of score-based diffusion models to learn task-agnostic behavior from offline, reward-free play datasets, without requiring hierarchical structures or additional clustering.
In addition, we demonstrated the effectiveness of Classifier-Free Guidance for simultaneously learning a goal-dependent and goal-independent policy in a sequential setting.
Experiments on several GCIL benchmarks showed that BESO significantly improves upon several state-of-the-art GCIL algorithms. 
Our ablation studies have demonstrated the key components of BESO that enable fast, deterministic behavior generation.
It further outperformed standard DDPM policies with only 3 denoising steps, alleviating prior drawbacks of slow diffusion sampling.

While BESO demonstrates great performance as a standalone policy, it also offers the flexibility to be seamlessly integrated into other hierarchical frameworks as an action prediction policy. 
Serving as a practical alternative to traditional behavior cloning policies, BESO sets itself apart with distinct features that are inherent to diffusion models.
In the future, we aim to extend BESO for language-guided behavior generation, offering more intuitive goal guidance for humans.

\section{Acknowledgments}

The work presented here was funded by the German Research Foundation (DFG) – 448648559.

\bibliographystyle{plainnat}
\bibliography{references}
\clearpage

\appendix

\subsection{BESO Hyperparameters}

\begin{table}
        \centering
        \begin{tabular}{l|ccc}  
        \toprule
         Hyperparameter & Block-Push & Relay-Kitchen & CALVIN   \\
             \midrule
                      Hidden Dimension       & 240 & 360 & 768 \\
                      Hidden Layers          &  4 & 6 & 6 \\
                      Window size            & 5 & 5 & 2 \\
                      Goal window size       & 1 & 2 & 1 \\
                      Number Heads           & 12 & 6 & 4 \\
                      Attention Dropout      & 0.05 & 0.3 & 0.2 \\
                      Residual Dropout       & 0.05 & 0 & 0.1 \\
                      Learning rate          & 1e-4 & 1e-4 & 1e-4 \\
                      Optimizer              & Adam & Adam & AdamW \\
                      Denoising steps        & 3 & 3 & 5 \\
                      $\sigma_{\text{max}}$  & 1 & 1 & 1 \\
                      $\sigma_{\text{min}}$ & 0.05   & 0.005 & 0.005 \\
                      $\sigma_{\text{data}}$  & 0.5  & 0.5 & 0.5 \\
                      best $\lambda$ for CFG & 2 & 1.25 & 1.25 \\
                      Noise distribution  & Log-Logistic & Log-Logistic &  Log-Logistic \\
                      EMA & True & True & True \\
                      Vision Encoder & None & None & ResNet-18 \\
                      Sampler Type & DDIM & DDIM & DDIM \\
                      Noise scheduler & Exp & Exp & Exp \\
                      Batch Size & 1024 & 1024 & 64 \\
                      Train steps in thousand & 60 & 40 & 120 \\
                                              
            \bottomrule
        \end{tabular}
        \caption{Overview of the most important Hyperparameters for the different model architectures for all tested environments.
        }
        \label{tab: Hyperparameter Overview simulation}
\end{table}
\label{sec: BESO Hyperparameters}
A summary of key hyperparameters of BESO is listed in Table \ref{tab: Hyperparameter Overview simulation}.
We observe, that transformer specific-hyperparameters such as the dropout rates require tuning according to the task,
while general diffusion hyperparameters remain consistent across different tasks. 

\textbf{Preconditioning.}
We utilize the preconditioning functions proposed in \citet{karras2022elucidating}: 
\begin{itemize}
    \item $c_{\text{skip}}= \sigma_{\text{data}}^2 / (\sigma_{\text{data}}^2 + \sigma_t^2)$
    \item $c_{\text{out}} = \sigma_t \sigma_{\text{data}} / \sqrt{\sigma_{\text{data}}^2 + \sigma_t^2}$
    \item $c_{\text{in}} = 1/ \sqrt{\sigma_{\text{data}}^2 + \sigma_t^2}$
    \item $ c_{\text{noise}} = 0.25 \ln (\sigma_t)$
\end{itemize}

\textbf{Normalization.}
BESO performs optimally when actions are diffused within a range of $[-1, 1]$ with a noise range of ${0.005, 1}$.
We adopted this noise range for all three environments, scaling the action output accordingly. 
For action diffusion with a larger range of, such as $[-3, 3]$, it is advisable to expand the noise range to higher values: $\{0.4, 40\}$ for optimal performance. 
For the input, we recommend normalizing the data with a mean of $0$ and a standard deviation of $1$. 

\textbf{Training Noise distribution.}
During training, noise values are sampled from a predefined noise distribution $P(\sigma)$. 
The standard distribution used in diffusion literature \cite{karras2022elucidating} is the log-normal, introducing two additional hyperparameters $\sigma_{\text{std}}, \sigma_{\text{max}}$, that require additional tuning. 
Our experiments revealed that the recommended values from prior work \cite{karras2022elucidating} are not optimal for action diffusion. 
Hence, we opted for the log-logistic distribution $\text{LogLogistic}(\alpha=0.5, \beta=0.5)$, which does not require additional parameters and works well in all our experiments. 

\textbf{Optimization.}
For optimization, we employed the commonly used Adam or AdamW optimizer for our experiments with a standard learning rate of $1e-4$.
Additionally, we use the Exponential Moving Average (EMA) to optimize our model's weights.

\begin{table}
    \centering
    \tiny
    \begin{tabular}{l|cc|cc}
      & \multicolumn{2}{c}{Block-Push} & \multicolumn{2}{c}{Relay Kitchen} \\
     & Reward & Result & Reward & Result \\
    \midrule
     Exponential & 0.96 ($\pm$ 0.02) & 0.93 ($\pm$ 0.02) & 3.86 ($\pm$ 0.09) & 3.74 ($\pm$ 0.09) \\
     Linear & 0.97 ($\pm$ 0.01) & 0.94 ($\pm$ 0.01) & 3.98 ($\pm$ 0.08) & 3.76 ($\pm$ 0.10) \\
     IDDPM \cite{nichol2021improved} & 0.95 ($\pm$ 0.01) & 0.92 ($\pm$ 0.01) & 3.88 ($\pm$ 0.08) & 3.67 ($\pm$ 0.09) \\
     Karras \cite{karras2022elucidating} & 0.96 ($\pm$ 0.02) & 0.93 ($\pm$ 0.03) & 3.96 ($\pm$ 0.08) & 3.75 ($\pm$ 0.07) \\
     VE \cite{song2020score}& 0.95 ($\pm$ 0.01) & 0.93 ($\pm$ 0.01) & 3.98 ($\pm$ 0.09)&  3.75 ($\pm$ 0.10) \\
     VP  \cite{song2020score}& 0.64 ($\pm$ 0.17) & 0.61 ($\pm$ 0.16) & 3.21 ($\pm$ 0.12) & 2.95 ($\pm$ 0.14) \\
    \bottomrule
    \end{tabular}
    \caption{Evaluation of the Influence of the time steps function for Goal-conditional Behavior Generation averaged over 10 seeds and 100 rollouts each. All models use the DDIM solver with 3 denosing steps for the comparison.  
    }
    \label{tab: noise scheduler ablation}
\end{table}
\textbf{Time steps.}
One important choice is the function of time steps, which determines how noise levels are distributed over the discrete steps.
Our empirical evaluation summarized in Table \ref{tab: noise scheduler ablation} indicates that exponential time steps are the most effective for BESO on average. 
However, other discretization methods such as the linear scheduler \cite{song2020score} and Karras scheduler \cite{karras2022elucidating} also deliver comparable results and can increase the performance on individual tasks.

\textbf{Recommendations.}
We recommend starting with the noise range of $\{ 0.005, 1 \}$ for a new task together with exponential time steps and the DDIM solver. To get the best performance, it is worth trying out other samplers such as Euler Ancestral and the linear time steps.

\subsection{Sampler Ablation}
\label{sec: Sampler Ablation studies}
We evaluate various state-of-the-art ODE samplers and their SDE counterparts in different environments.
To determine the best solver for conditional-behavior generation, we analyze the average performance of 10 different seeds with 100 rollouts each in different environments. 
In general, we differentiate first-order and second-order solvers: the first order solver is Euler \cite{ho2020denoising} and the tested second order solver is Heun \cite{karras2022elucidating}. 
The tested samplers include:
\begin{itemize}
    \item \textbf{Euler ODE (Euler):}  A first-order ODE sampler from \cite{karras2022elucidating} without the additional addition and deleting of noise. 
    The algorithm is summarized in \ref{alg: Deterministic 1st order sampler}.
    \item \textbf{Euler-Ancestral (EA):} A continuous-time version of the standard DDPM sampler  \cite{ho2020denoising} introduced in \cite{song2020score}. 
    \item \textbf{2nd Order Heun Solver (Heun):} A second-order ODE solver using the Heun method \cite{karras2022elucidating}. 
    \item \textbf{DPM:} An exponential ODE integrator solver designed for synthesis in a few inference steps \cite{lu2022dpmsolver}. We use the second order method.
    \item \textbf{DDIM:} A first order variant of DPM, which has been introduced individually \cite{song2021denoising, lu2022dpmsolver} and has been designed for fast inference and CFG.
    \item \textbf{DPM-Ancestral:}A stochastic variant of DPM with ancestral noise injections.  
    \item \textbf{DPM++(2S):}  An improved version of the second order DPM sampler for classifier-free guidance based conditional diffusion models with a single inference step \cite{lu2022dpm}
    \item \textbf{DPM++(2M):}  An improved version of the second-order DPM sampler for classifier-free guidance based conditional diffusion models \cite{lu2022dpm}, which is a second order method using two model predictions per step.
\end{itemize}
Several previous studies have compared the performance of ODE samplers in the context of image generation \cite{karras2022elucidating, lu2022dpm}. 
However, these comparisons may not be entirely indicative as image generation tasks have unique challenges and requirements not relevant to action synthesis.
To ensure a fair comparison, we evaluated all samplers on the same models across several simulation environments and report their average performance based on 100 runs for each environment. 
This allows us to accurately compare the effectiveness of each deterministic solver in the context of step-based action generation.
The results for the kitchen environment are shown in Table \ref{tab: deterministic kitchen sampler study} and the performance for the block push is reported in Table \ref{tab: deterministic push sampler study}.
As shown in both tables, the first-order exponential integrator solver DDIM achieves the best overall performance.
Increasing the number of inference steps does not have a significant impact on the average performance, even reducing the average result of some samplers.
Overall the performance differences of all evaluated samplers are small.

\subsection{Baselines Implementation}
\label{sec: baseline implementation}
The MLP-based models have $4$ layers with $512$ neurons and use the ReLU activation function.
All diffusion models have the same transformer backbone, and C-BeT uses its recommended parameters.
During training, the Adam optimizer was used with a learning rate of $0.001$ for MLP models and $1e-4$ for transformer models.  
The batch size for MLP models was $512$, while it was $1024$ for transformer models, except for BeT, which used a batch size of $64$ as recommended in \cite{cui2023from}.

\textbf{GCBC}
For the GCBC model, the goal is concatenated with the state and fed into the 4-layer MLP architecture with a dropout rate of 0.1.

\textbf{GC-IBC}
The GC-IBC model uses the same MLP architecture as GCBC and is optimized using the InfoNCE loss with additional energy-regularization and Wasserstein Gradient loss. During experiments, adding a penalty term with $\lambda=0.005$ to restrict the average energy improved training stability \cite{florence2022implicit}.
Given the large number of tunable hyperparameters for IBC, we ran a hyperparameter search to determine the best ones. 
We want to note, that the model results of EBM were very sensitive to initial seeds and we had trouble getting consistent results for the models.
Similar observations of IBC performance have been reported in related work \cite{pearce2023imitating, chi2023diffusionpolicy}.

\begin{table}[]
    \centering
    \begin{tabular}{c|cc}
    \toprule
          Hyperparameter & Block Push & Relay Kitchen  \\
          \midrule
         Hidden dimension & 128 & 128 \\
          Hidden layers & 6 & 6 \\
          Train steps  & 5000 & 1000 \\
        Noise Scale  & 0.3 & 0.3 \\
         Loss & InfoNCE & InfoNCE \\
         Train samples & 64 & 64 \\
         Noise shrink & 0.5 & 0.5 \\
         Learning rate & 0.001 & 0.001 \\
        \bottomrule
    \end{tabular}
    \caption{Overview of the used Hyperparameters of GC-IBC for both environments.}
    \label{tab:my_label}
\end{table}

\textbf{C-BeT}
For the performance of C-BeT, we use the recommended parameters from \citet{cui2023from} for all tested environments. Our reported results are marginally worse, than the ones reported in the original work, since they do not average it over 10 seeds.

\textbf{Latent Motor Plans}
\begin{table}
\tiny
        \centering
        \begin{tabular}{l|ccc}  
        \toprule
         Hyperparameter & Block-Push & Relay-Kitchen & CALVIN   \\
             \midrule
          Decoder Hidden Dimension       & \{128, 256, 512, \textbf{1024}\} & \{128, 256, 512\} & 2048 \\
          n-Mixtures             & 10 & 10 & 10 \\
          n-Classes              &  \{10, 32, 64, 128, \textbf{256}\} & 10 & 10 \\
          Policy-dropout         &  \{0.1, \textbf{0.2}, 0.3\} & \{0.1, \textbf{0.2}, 0.3\} & 0.1 \\
          Plan Features          & \{16, \textbf{32}, 64, 128\} & \{16, 32, \textbf{64}, 128\} & 32 \\
          Plan Recognition Features   & \{64, 128, \textbf{256}, 512\} & \{64, 128, 256, \textbf{512}\} & 2048 \\
          Replan Freq            & \{5, \textbf{10}, 16, 32\} & \{5, 10, 16, \textbf{32} \} & 2 \\
          Planner Hidden Layers          & 2 & 2 & 2 \\
          Window size            & \{10, 16, 32, \textbf{48}\} & \{10, \textbf{16}, 32, 48, 64\} & 16 \\
          Goal window size       & 1 & 1 & 1 \\
          kl-beta                & \{0.001, 0.005, \textbf{0.01}\} & \{0.001, 0.005, \textbf{0.01}\} & 0.01 \\
          Learning rate          & \{0.001, \textbf{0.0005}, 0.0001\} &  \{0.001, \textbf{0.0005}, 0.0001\} & 0.0001 \\
          Optimizer              & Adam & Adam & Adam \\
            \bottomrule
        \end{tabular}
        \caption{Overview of the Hyperparameter-Sweep for Latent Plans and the final parameters used for the evaluation for each tested simulation environment
        }
        \label{tab:  LMP Hyperparameter Overview simulation}
\end{table}
The LMP model was evaluated on the Kitchen and Block Push environments with extensive hyper-parameter sweeps to find the best-performing configuration.
A detailed overview of the sweep parameters and the chosen ones is shown in Table \ref{tab:  LMP Hyperparameter Overview simulation}.
On the CALVIN environment, the proposed parameters from prior work were used \cite{rosete-beas2022latent}. 
We used the improved LMP variant, called HULC, from \cite{mees2022hulc}, which uses a different Kl-divergence weighting term and a transformer model the Seq2Seq CVAE.

\textbf{RIL}
For the low-level policy of kitchen and block push we use 4 layers with 512 neurons each. 
For the CALVIN task, we use the baseline version from \cite{rosete-beas2022latent} and kept the hyperparameters the same for training.

\textbf{Diffusion-X}
The baseline from \cite{pearce2023imitating} uses the same hyper-parameters of our transformer model reported in \ref{tab: Hyperparameter Overview simulation} to guarantee a fair comparison.
Diffusion-X uses $50$ inference steps on the kitchen task combined with additional 10 fine-tuning steps at the lowest noise level, while we use $20$ inference steps for the block-push environment and additional $8$ fine-tuning steps. 
Diffusion-X uses a discrete variant of the Euler sampling method with an ancestral noise scheduler, which is reported in Alg. \ref{alg: Ancestral Sampling} \cite{ho2020denoising, song2020score}. 
Further, it applies $X$-additional denoising steps at the lowest noise level.

\begin{algorithm}
\label{alg: Ancestral Sampling}
    \caption{Ancestral Noise Scheduler $f_{\text{ANC}}$ \cite{song2020score, ho2020denoising} }
    \begin{algorithmic}[1] 
    \State \textbf{Require:} $t_{\text{from}}$, $t_{\text{to}}$
        \State $t_{\text{up}} \gets \min( t_{\text{to}},  \sqrt{\frac{t_{\text{to}}^2 (t_{\text{to}}^2 - t_{\text{from}}^2)}{t_{\text{from}}^2}} )$
        \State $t_{\text{to}} \gets  \sqrt{(t_{\text{to}}^2 - t_{\text{from}}^2)} $
        \State{\textbf{return} $t_{\text{down}}$, $t_{\text{up}}$}
    \end{algorithmic}
\end{algorithm}

\begin{algorithm}
\label{alg: Deterministic 1st order sampler}
    \caption{Deterministic 1st Order Euler Sampler \cite{karras2022elucidating} }
    \begin{algorithmic}[1] 
    \State \textbf{Require:} Current state $\state$, goal $\mathbf{g}$
    \State \textbf{Require:} Score-Denoising Model $D_{\theta}(\mathbf{a}, \state,\mathbf{g}, \sigma)$
    \State \textbf{Require:} Noise scheduler $\sigma_t = \sigma(t_{i})$
    \State \textbf{Require:} Discrete time steps $t_{i \in \{ 0,.., N\}}$
        \State Draw sample $\mathbf{a}_0 \sim \mathcal{N}( \mathbf{0}, \sigma_{0}^2\mathbf{I})$
        \For{$i \in \{0, ...,N-1 \}$ } 
            \State{ $\mathbf{d}_{i} \gets \big( \mathbf{a}_i - D_{\theta}(\mathbf{a}_i, \state, \mathbf{g}, \sigma_{i}) \big) / \sigma_{i}$}
            \State {$\mathbf{a}_{i+1} \gets \mathbf{a}_{i} + (t_{i+1}- t_i) \mathbf{d}_i$}
        \EndFor
        \State{\textbf{return} $\mathbf{a}_N$}
    \end{algorithmic}
\end{algorithm}

\begin{algorithm}
\label{alg: stochastic 1st order sampler}
    \caption{Stochastic 1st Order Euler sampler \cite{karras2022elucidating} }
    \begin{algorithmic}[1] 
    \State \textbf{Require:} Current state $\state$, goal $\mathbf{g}$
    \State \textbf{Require:} Score-Denoising Model $D_{\theta}(\mathbf{a}, \state,\mathbf{g}, \sigma)$
    \State \textbf{Require:} Noise scheduler $\sigma_i = t_{i}$, $f_{\text{ANC}}$ from Alg. \ref{alg: Ancestral Sampling}
    \State \textbf{Require:} Discrete time steps $t_{i \in \{ 0,.., N\}}$
        \State Draw sample $\mathbf{a}_0 \sim \mathcal{N}( \mathbf{0}, \sigma_{0}^2\mathbf{I})$
        \For{$i \in \{0, ...,N-1 \}$ } 
            \State{$\mathbf{d}_{i} \gets \big( \mathbf{a}_i - D_{\theta}(\mathbf{a}_i, \state, \mathbf{g}, \sigma_{i}) \big) / \sigma_{i}$}
            \State {$t_{\text{down}}, t_{\text{up}} \gets f_{\text{ANC}}(t_i, t_{i+1})$}
            \State {$\mathbf{a}_{i+1} \gets \mathbf{a}_{i} + (t_{\text{down}} - t_i) \mathbf{d}_i$}
            \State $\epsilon_{\text{up}} \sim  \mathcal{N}( \mathbf{0}, \sigma_{t_{\text{up}}}^2 \mathbf{I}) $
            \State $\mathbf{a}_{i+1} \gets \mathbf{a}_{i+1} + \epsilon_{\text{up}}$
        \EndFor
        \State{\textbf{return} $\mathbf{a}_N$}
    \end{algorithmic}
\end{algorithm}

\begin{table*}
        \centering
        \begin{tabular}{lc|cccccc}  
        
       & Steps & Euler  &  Heun & DDIM &  DPM  & DPM++(2S)  & DPM++(2M)  \\
       \toprule
     \multirow{5}{*}{Reward} & 3   & 3.87 ($\pm$ 0.09) & 3.80 ($\pm$ 0.03) & 3.92 ($\pm$ 0.07) & 3.86 ($\pm$ 0.08) & 3.88 ($\pm$ 0.08) & 3.89 ($\pm$ 0.08) \\
                             & 5   & 3.87 ($\pm$ 0.07) & 3.84 ($\pm$ 0.06) & 3.88 ($\pm$ 0.06) & 3.90 ($\pm$ 0.09) & 3.87 ($\pm$ 0.06) & 3.87 ($\pm$ 0.06) \\
                             & 10  & 3.85 ($\pm$ 0.08) & 3.86 ($\pm$ 0.09) &  3.88 ($\pm$ 0.06) & 3.87 ($\pm$ 0.10) & 3.88 ($\pm$ 0.07) & 3.89 ($\pm$ 0.05)  \\
                             & 20  & 3.86 ($\pm$ 0.10) & 3.91 ($\pm$ 0.08) & 3.87 ($\pm$ 0.07) & 3.88 ($\pm$ 0.09) & 3.89 ($\pm$ 0.07) & 3.88 ($\pm$ 0.06)   \\ 
                             & 50  & 3.82 ($\pm$ 0.08) & 3.93 ($\pm$ 0.04) & 3.88 ($\pm$ 0.06) & 3.82 ($\pm$ 0.10) & 3.67 ($\pm$ 0.04) & 3.89 ($\pm$ 0.06) \\
             \midrule
   \multirow{5}{*}{Result}   & 3  & 3.66 ($\pm$ 0.09) & 3.62 ($\pm$ 0.07) & 3.69 ($\pm$ 0.07) & 3.67 ($\pm$ 0.10) & 3.67 ($\pm$ 0.09) & 3.67 ($\pm$ 0.08)   \\
                             & 5  & 3.66 ($\pm$ 0.07) & 3.66 ($\pm$ 0.06) & 3.67 ($\pm$ 0.08)& 3.67 ($\pm$ 0.08) & 3.66 ($\pm$ 0.07) & 3.66 ($\pm$ 0.08) \\
                             & 10 & 3.65 ($\pm$ 0.06) & 3.63 ($\pm$ 0.06) & 3.67 ($\pm$ 0.07) & 3.66 ($\pm$ 0.07) & 3.66 ($\pm$ 0.08) & 3.67 ($\pm$ 0.07)  \\
                             & 20 & 3.64 ($\pm$ 0.07) & 3.65 ($\pm$ 0.09) & 3.66 ($\pm$ 0.09) & 3.66 ($\pm$ 0.07) & 3.68 ($\pm$ 0.09) & 3.67 ($\pm$ 0.08)  \\ 
                             & 50 & 3.62 ($\pm$ 0.04)  & 3.67 ($\pm$ 0.04) & 3.67 ($\pm$ 0.09) & 3.62 ($\pm$ 0.07) & 3.67 ($\pm$ 0.07) & 3.67 ($\pm$ 0.08) \\
            \bottomrule
        \end{tabular}
        \caption{Comparison of the performance of deterministic samplers on the Kitchen Environment averaged over 10 seeds with 100 rollouts each. We tested BESO trained on log-normal noise distribution with $\sigma_{\text{mean}}=-2$, $\sigma_{\text{std}}=-2$, $\sigma_{\text{max}}=33$, $\sigma_{\text{min}}=0.39$ and use the exponential time steps.
        }
        \label{tab: deterministic kitchen sampler study}
\end{table*}

\begin{table*}
        \centering
        \begin{tabular}{lc|cccccc}  
        
       & Steps & Euler  &  Heun & DDIM &  DPM  & DPM++(2S)  & DPM++(2M)  \\
       \toprule
     \multirow{5}{*}{Reward} & 3 & 0.95 ($\pm$ 0.02)  & 0.92 ($\pm$ 0.02) & 0.96 ($\pm$ 0.02) & 0.96 ($\pm$ 0.02)  & 0.95 ($\pm$ 0.03) & 0.97 ($\pm$ 0.02)\\
                             & 5   & 0.94 ($\pm$ 0.04) & 0.95 ($\pm$ 0.02) & 0.96  ($\pm$ 0.02) & 0.97 ($\pm$ 0.01) & 0.94 ($\pm$ 0.02) & 0.93 ($\pm$ 0.02) \\
                             & 10  & 0.97 ($\pm$ 0.03) & 0.93 ($\pm$ 0.02) &  0.96 ($\pm$ 0.01) & 0.95 ($\pm$ 0.03) & 0.96 ($\pm$ 0.02) & 0.96 ($\pm$ 0.03)  \\
                             & 20  & 0.98 ($\pm$ 0.02) & 0.96 ($\pm$ 0.03) & 0.98 ($\pm$ 0.02) & 0.96 ($\pm$ 0.03) & 0.96 ($\pm$ 0.02) & 0.97 ($\pm$ 0.03)   \\ 
                             & 50 & 0.98 ($\pm$ 0.01)  & 0.96 ($\pm$ 0.01) & 0.97 ($\pm$ 0.02) & 0.97 ($\pm$ 0.05) & 0.97 ($\pm$ 0.01) & 0.94 ($\pm$ 0.05) \\
             \midrule
   \multirow{5}{*}{Result}   & 3 & 0.94 ($\pm$ 0.02)  & 0.90 ($\pm$ 0.05) & 0.94 ($\pm$ 0.04) &  0.94 ($\pm$ 0.01) & 0.92 ($\pm$ 0.03) & 0.94 ($\pm$ 0.03)\\
                             & 5  & 0.91 ($\pm$ 0.06) & 0.93 ($\pm$ 0.03) & 0.95 ($\pm$ 0.02)& 0.95 ($\pm$ 0.02) & 0.91 ($\pm$ 0.03) & 0.93 ($\pm$ 0.03) \\
                             & 10 & 0.94 ($\pm$ 0.02) & 0.91 ($\pm$ 0.04) & 0.95 ($\pm$ 0.02) & 0.91 ($\pm$ 0.04) & 0.94 ($\pm$ 0.02) & 0.96 ($\pm$ 0.02)  \\
                             & 20 & 0.96 ($\pm$ 0.02) & 0.94 ($\pm$ 0.03) & 0.95 ($\pm$ 0.04) & 0.95 ($\pm$ 0.04) & 0.93 ($\pm$ 0.03) & 0.96 ($\pm$ 0.03)  \\ 
                             & 50 & 0.98 ($\pm$ 0.01)  & 0.95 ($\pm$ 0.02) & 0.95 ($\pm$ 0.01) & 0.93 ($\pm$ 0.03) & 0.94 ($\pm$ 0.03) & 0.92 ($\pm$ 0.06) \\
            \bottomrule
        \end{tabular}
        \caption{Comparison of the performance of deterministic samplers on the Block Push environment averaged over 10 seeds with 100 rollouts each. We tested BESO trained on log-normal noise distribution with $\sigma_{\text{mean}}=-0.17$, $\sigma_{\text{std}}=-2$, $\sigma_{\text{max}}=40.5$, $\sigma_{\text{min}}=0.39$ and use the exponential time steps.
        }
        \label{tab: deterministic push sampler study}
\end{table*}

\end{document}